\documentclass[runningheads]{llncs}
\usepackage[T1]{fontenc}
\usepackage{cite}
\usepackage{amsmath,amssymb,amsfonts}
\usepackage{algorithm}
\usepackage{algorithmic}
\usepackage{arevmath} 
\usepackage[margin=2cm]{geometry}
\usepackage{graphicx}
\usepackage{colortbl}
\usepackage{textcomp}
\usepackage{xcolor}
\def\BibTeX{{\rm B\kern-.05em{\sc i\kern-.025em b}\kern-.08em
    T\kern-.1667em\lower.7ex\hbox{E}\kern-.125emX}}

\def\ie{\emph{i.e.}}
\def\etc{\emph{etc.}}

\definecolor{myblue}{HTML}{55caff}
\definecolor{mygreen}{HTML}{55d555}

\definecolor{greenF2}{HTML}{00c41e}
\definecolor{redF2}{HTML}{FF5959}
\definecolor{blueF2}{HTML}{7b84f7}
\definecolor{orangeF2}{HTML}{ffa000}
\definecolor{magentaF2}{HTML}{FF7FFF}
\definecolor{yellowF2}{HTML}{ffdd37}
\definecolor{oilF2}{HTML}{d2dd37}
\definecolor{purpleF2}{HTML}{E0B0FF}
\definecolor{cyanF2}{HTML}{00d4f7}
\definecolor{green2F2}{HTML}{00e884}

\begin{document}

\newcommand\myeq{\mkern1.5mu{=}\mkern1.5mu}

\newcommand{\gfc}[1]{{\color{cyan}~{[\small\bf Germain:} #1{\bf ]}}}

\title{Finding Foundation Models for Time Series Classification\\ with a PreText Task}


\author{Ali Ismail-Fawaz\inst{1} \and
Maxime Devanne\inst{1} \and
Stefano Berretti\inst{2} \and Jonathan Weber\inst{1}  \and \\ Germain Forestier \inst{1,3}}

\authorrunning{A. Ismail-Fawaz et al.}

\institute{IRIMAS, Université de Haute-Alsace, France
\email{\{ali-el-hadi.ismail-fawaz,maxime.devanne,jonathan.weber,germain.forestier@uha.fr\}@uha.fr} \and
MICC, University of Florence, Italy \email{stefano.berretti@unifi.it} \and
DSAI, Monash University, Australia \email{germain.forestier@monash.edu}}

\maketitle

\begin{abstract}

Over the past decade, Time Series Classification (TSC) has gained an increasing attention.
While various methods were explored, deep learning -- particularly through Convolutional Neural Networks (CNNs) --stands out as an effective approach.
However, due to the limited availability of training data, defining a foundation model for TSC that overcomes the overfitting problem is still a challenging task.
The UCR archive, encompassing a wide spectrum of datasets ranging from motion recognition to ECG-based heart disease detection, serves as a prime example for exploring this issue in diverse TSC scenarios.
In this paper, we address the overfitting challenge by introducing pre-trained domain foundation models.
A key aspect of our methodology is a novel pretext task that spans multiple datasets. This task is designed to identify the originating dataset of each time series sample, with the goal of creating flexible convolution filters that can be applied across different datasets.
The research process consists of two phases: a pre-training phase where the model acquires general features through the pretext task, and a subsequent fine-tuning phase for specific dataset classifications.
Our extensive experiments on the UCR archive demonstrate that this pre-training strategy significantly outperforms the conventional training approach without pre-training.
This strategy effectively reduces overfitting in small datasets and provides an efficient route for adapting these models to new datasets, thus advancing the capabilities of deep learning in TSC.

\end{abstract}


\section{Introduction}

Time series are sequences of data points indexed by time, typically obtained by observing a random variable over consistent intervals. 
These data sequences are prevalent in various machine learning applications, including classification~\cite{middlehurst2023bake}, clustering~\cite{ismail-fawaz2023shapedba}, and extrinsic regression~\cite{guijo2023unsupervised}, among others. 
Over the past decade, Time Series Classification (TSC) has witnessed a surge in research activity. This increasing interest spans across diverse fields such as medicine and telecommunications.

Deep learning, with its advanced neural network architectures, offers significant potential for Time Series Classification (TSC)~\cite{ismail2019deep}, often achieving state-of-the-art performance in various TSC tasks.
Conventionally, solving a TSC problem with deep learning involves initializing a neural network architecture randomly and fitting it with the training data. 
However, when the training dataset is limited, this method can lead to overfitting, where the model adapts too closely to the training data, resulting in poor performance on unseen test samples.
This challenging problem of having a dataset with few training examples does exist almost everywhere in machine learning research.
This common problem reflects a real case scenario and it has been adapted to datasets of the UCR archive, the most comprehensive repository for univariate TSC datasets.

\begin{figure}
    \centering
    \includegraphics[width=0.5\linewidth]{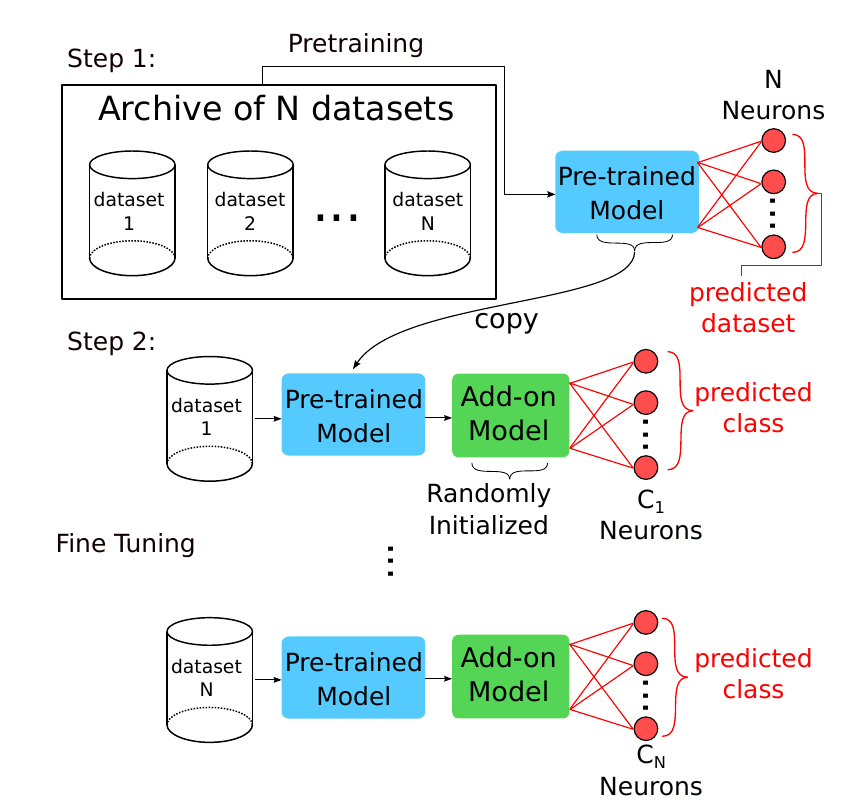}
    \caption{Summary of the proposed pretext task approach.
    Given an archive of $N$ datasets, the first step is to train a \emph{pre-trained} model (in \colorbox{myblue}{blue}) on all of the datasets, where the classification task is to predict the dataset each time series belongs to.
    The second step is to copy the \emph{pre-trained} model and follow it with an \emph{addon} model (in \colorbox{mygreen}{green}) randomly initialized.
    The second step is done for each of the $N$ datasets of the archive independently.
    After constructing the $N$ new models, they are fine tuned on each dataset depending on the task of each one.}
    \label{fig:summary}
\end{figure}

This large archive is composed of 128 datasets of TSC covering various tasks going from motion recognition to the classification of heart diseases using Electrocardiogram (ECG) signals.
The UCR archive's depth lies in its diverse representation of tasks across multiple domains, often providing several example datasets for each domain.

Gathering additional training samples to address the overfitting issue can be time-consuming and resource-intensive.
Furthermore, even if more samples are generated, annotating them typically necessitates expertise, thus introducing additional costs. 
As a solution, various approaches were proposed in the literature
such as data augmentation~\cite{pialla2022data,IsmailFawaz2018Dataaugmentationusing}, and the use of hand-crafted generic filters~\cite{ismail-fawaz2022hccf}.
However, while effective, these methods can sometimes introduce noise and disrupt the training process.

To take advantage of having multiple datasets within a given domain, we aim to identify a foundation pre-trained model for each domain of TSC
, replacing the random initialization used in traditional techniques.
This
\emph{pre-trained} foundation model is trained on a shared task among the different datasets. Specifically, the task is to predict the original dataset of each sample.
For instance, if we merge two datasets, \textit{dataset1} and \textit{dataset2}, from the same domain and temporarily disregard their specific target classes, the pre-trained model's objective becomes discerning the origin of each sample in this combined set.

Once the pre-training phase is completed, the model is fine-tuned for the specific tasks of each dataset. A concise overview of our proposed methodology is depicted in Figure~\ref{fig:summary}.

After the pre-trained model has been fully trained on the pretext task, the fine tuning stage can follow two different options.
The first option is to fine tune the pre-trained model followed by a classification layer with respect to the dataset's classification task.
The second option is to fine tune the pre-trained model cascaded with deeper layers to extract deeper features followed by a classification layer.
The first option was used in the work of~\cite{fawaz2018transfer}, where the authors studied the effect of transfer learning on TSC but performance was not as good as expected. 
In other words, most target datasets were sensitive on the dataset used as source for the transfer learning.
In this work follow the setup of the second option.
The main reason we believe the first option may cause issues, is the fact that ignoring deeper meaningful features correlated with one dataset during the fine tuning step implies a strong assumption: the pre-trained model learned the optimal convolution filters that are able to correctly generalize to the classification task.
However, because this may not be the case, in this work we decided to follow the second option.



The main contributions of this work are:
\begin{itemize}
    \item Novel domain foundation models trained to solve a pretext task to enhance deep learning for TSC;
    \item  Novel Batch Normalization Multiplexer (BNM) layer that con- trols the multi-dataset (multi-distribution) problem of the batch normalization;
    \item Extensive experiments on the UCR archive show a significant improvement when using the pre-trained model over the baseline model;
    \item Extensive experiments on the UCR archive show a significant improvement when using the pre-trained model over the baseline model.
\end{itemize}

\section{Related Work}
Many works in the literature have been proposed to address the TSC task and have been evaluated on the UCR archive.
These tasks range from similarity based approaches to ensemble models, deep learning, \etc
In what follows, we present the latest state-of-the-art approaches that addressed the TSC task.

\subsection{Non Deep Learning Techniques}
A basic approach for solving this type of classification task is simply by using the Nearest Neighbor algorithm.
To use it on time series data, a distance function should be defined, such as the Dynamic Time Warping (DTW) measure.
DTW is a more suitable measure to be used compared to the Euclidean distance, which is traditionally in use.
The usage of DTW is seen to be better than the Euclidean due to its ability to align the time series before measuring the distance between them.
Coupled with DTW, the Nearest Neighbor algorithm is seen to be very effective~\cite{bagnall2017great,lines2015time}, and was set to be the baseline to new TSC approaches.
This gave rise to the definition of a barycenter of time series examples, and to their use for tasks such as NN to solve TSC~\cite{petitjean2014dynamic}.

Given this TSC base approach, more algorithms appeared in the literature that showed a significant improvement in performance. 
For instance, the authors of~\cite{middlehurst2021hive} proposed an ensemble, called HIVE-COTE2.0 (HC2) of multiple TSC machine learning algorithms.
Although HC2 is powerful compared to other approaches, it still has the problem of training time, which can last for days.
For this reason, the authors of~\cite{dempster2020rocket} proposed a random convolution based model that is faster than all state-of-the-art approaches and was enhanced in~\cite{dempster2021minirocket,tan2022multirocket} with its latest version MultiROCKET.
MultiROCKET achieves state-of-the-art performance with a small training time.

Finally, some approaches addressed the TSC task by using a dictionary based model.
For instance, the authors of~\cite{dempster2023hydra} proposed a dictionary based method based on ROCKET's ideology.
Coupled with MultiROCKET, it can achieve better results than MultiROCKET alone~\cite{middlehurst2023bake}.
More recently, the authors in~\cite{schafer2023weasel} proposed WEASEL2.0, the new version of WEASEL~\cite{schafer2017fast}, which is a sliding window approach for transforming the time series into feature vectors.

\subsection{Deep Learning Techniques}
In 2019, the authors of~\cite{ismail2019deep} released a detailed review on the latest deep learning approaches for solving TSC on the UCR archive.
The two best performing models were Convolutional Neural Networks (CNNs), the Fully Convolutional Network (FCN) and the Residual Network (ResNet)~\cite{wang2017time}.
Moreover, the authors of~\cite{ismail2020inceptiontime} proposed a new CNN based architecture called InceptionTime, which is an ensemble of multiple Inception models.
More recently, new hand-crafted convolution filters were proposed to enhance InceptionTime by~\cite{ismail-fawaz2022hccf} with their proposed model H-InceptionTime achieves new state-of-the-art performance for deep learners on TSC.
Finally, the authors of~\cite{Ismail-Fawaz2023LITELightInception} argued that there is no need for large complex models to solve the TSC task on the UCR archive, but instead they proposed a lighter architecture called LITE. 
LITE balances between its small number of parameters and its state-of-the-art performance using some boosting techniques.

\subsection{Pre-Training Deep Learning Techniques}
In the last few years, some approaches addressed the TSC task using pre-trained deep learning models.
For instance, the work in~\cite{fawaz2018transfer} proposed to apply transfer learning of a deep learning model from a source time series dataset to a target dataset.
In other words, the deep learning model was trained on a source dataset and then fine tuned on a target dataset.
Moreover, some work consisted on training a deep learning model with a Self-Supervised task and then use its output features to learn a classifier~\cite{ismail-fawaz2022trilite}.
Another technique in using pre-trained models is the so called ``knowledge distillation'', where the authors of~\cite{ay2022study} used a pre-trained FCN~\cite{wang2017time} model and distilled its knowledge to a smaller version of FCN. 
This process helps to balance between a smaller architecture and its performance.
In~\cite{xu2023distilling}, authors addressed as well the task of TSC by distilling knowledge from a pre-trained model using an adversarial approach that discriminates data domain.

The difference between our proposed approach and the traditional pre-training techniques is the usage of multiple domains during training.
It is important to note that the goal of this work is not to solve transfer learning but instead to enhance deep learners when solving direct TSC tasks using a pre-training approach. 
In what follows, we detail our approach and the pretext task used.

\section{Proposed Method}
First, we introduce some definitions that will be used in the subsequent sections of this work.

\subsection{Definitions}
\begin{itemize}
    \item A Multivariate Time Series (MTS) $\textbf{X} = \{\textbf{x}_0, \textbf{x}_1, \ldots, \textbf{x}_d\}$ is a set of $d$ Univariate Time Series.
    \item A Univariate Time Series (UTS) $\textbf{x} = \{x_0, x_1, \ldots, x_T\}$ is a vector of $T$ values of a random variable changing with time.
    \item Univariate Time Series Classification Dataset (UTSCD) $\mathcal{D}=\{(\textbf{x}_i,\textbf{y}_i)\}_{i=1}^{N-1}$ is a set of $N$ UTS with their corresponding label vector $\textbf{y}$. We denote by $C$ the number of unique labels existing in $D$.
\end{itemize}

\begin{algorithm}
\caption{Train the Pre-Trained Model on pretext Task}
\label{alg:pre-trained}
\begin{algorithmic}[1]
 \renewcommand{\algorithmicrequire}{\textbf{Input:}}
 \renewcommand{\algorithmicensure}{\textbf{Output:}}
\REQUIRE
    $\mathcal{D}=\{\mathcal{D}_1,\mathcal{D}_2\ldots\mathcal{D}_N\}$ N datasets of UTSC where $\mathcal{D}_i=\{\textbf{x}_{ij},y_{ij}\}_{j=1}^{M_i}$, the number of layers for the pre-trained mode $L_{PT}$
\ENSURE A pre-trained model $PT(.)$ trained on the pretext task over all the datasets in $\mathcal{D}$

\STATE Define $M = sum(M_1,M_2,\ldots,M_{N})$
\STATE Define $\mathcal{D}_{PT} = empty List$
\STATE Build $PT(.)$ a neural network with $L_{PT}$ layers and $M$ output units with $softmax$ activation
\FOR{$i=1$ to $N$}
    \FOR{$j=1$ to $M_i$}
        \STATE $\mathcal{D}_{PT}.append([\textbf{x}_{ij},i])$
    \ENDFOR
\ENDFOR
\STATE $PT.train(\mathcal{D}_{PT})$
\RETURN $PT(.)$

\end{algorithmic}
\end{algorithm}

\begin{algorithm}
\caption{Fine Tuning on Each Dataset}
\label{alg:finetune}
\begin{algorithmic}[1]
\renewcommand{\algorithmicrequire}{\textbf{Input:}}
\renewcommand{\algorithmicensure}{\textbf{Output:}}
\REQUIRE $\mathcal{D}=\{\mathcal{D}_1,\mathcal{D}_2\ldots\mathcal{D}_N\}$ N datasets of UTSC where $\mathcal{D}_i=\{\textbf{x}_{ij},y_{ij}\}_{j=1}^{M_i}$, a pre-trained model $PT(.)$ of $L_{PT}$ layers trained on the pretext task, the number of layers of an addon model while fine tuning $L_{FT}$
\ENSURE $\{FT_1(.),FT_2(.),\ldots FT_N(.)\}$ $N$ fine tuned models of $L_{PT}+L_{FT}$ layers trained on the task of each dataset independently

\STATE Build $\{FT_1(.),FT_2(.),\ldots, FT_N(.)\}$ neural networks of $L_{PT}+L_{FT}$ layers with output nodes respecting the number of classes of each dataset in $\mathcal{D}$ respetively
\STATE Fill the first $L_{PT}$ layers in $\{FT_1(.),FT_2(.),\ldots,FT_N(.)\}$ by the feature extraction part of $PT(.)$
\FOR{$i=1$ to $N$}
    \STATE $FT_i.train(\mathcal{D}_i)$
\ENDFOR

\RETURN $\{FT_1(.),FT_2(.),\ldots, FT_N(.)\}$

\end{algorithmic}
\end{algorithm}

\subsection{Pretext Task}
Given a backbone deep learning model for TSC made of $n$ layers, we divided the backbone model into two sub-models.
The first sub-model (referred to as the pre-trained model) focuses on learning a pretext task and the latter is an additional randomly initialized model acting as an add-on to the pre-trained model that focuses on the TSC task.
The pretext task chosen in this work is the following: given a set of $M$ UTSCD, the pre-trained model's task is to correctly predict from which dataset each sample belongs to (see Algorithm~\ref{alg:pre-trained}).
It is important to note that one could argue that a more intuitive approach is to combine all datasets and classes and predict a massive class distribution without the need of going through a pretext task.
This last approach, however, would result in some issues when no correlation exists between classes of different datasets, so that the class distribution would not have a meaningful representation.

\begin{figure*}
    \centering
    \includegraphics[width=\textwidth]{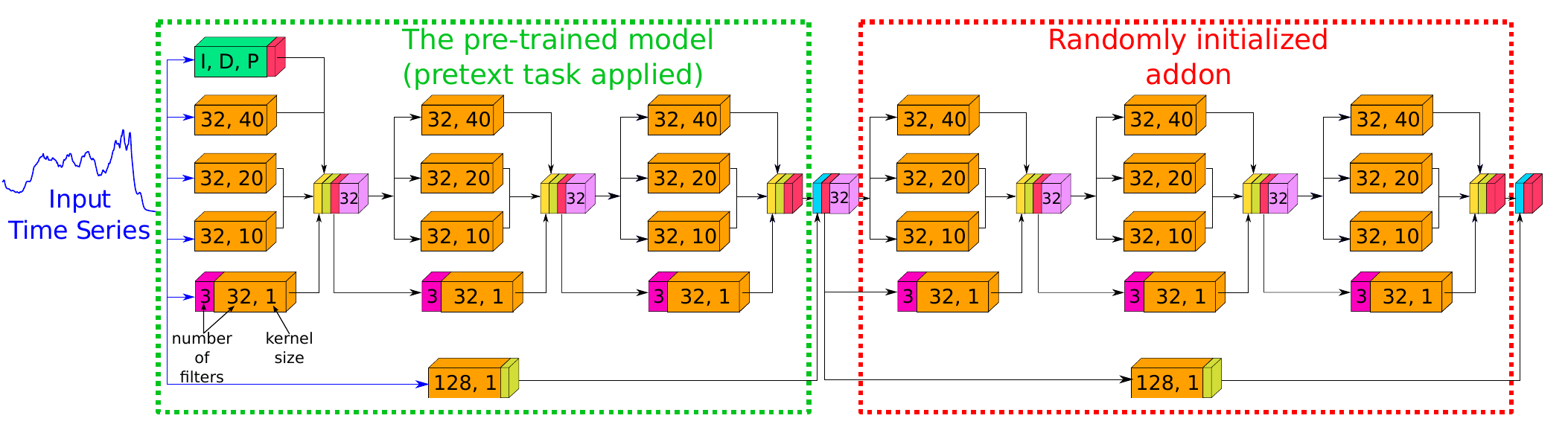}
    \caption{The architecture of H-Inception divided into two sub-models.
    The first model is the pre-trained model, trained on the pretext task (dotted \colorbox{greenF2}{green} rectangle), while the second model is the randomly initialized add-on model (dotted \colorbox{redF2}{red} rectangle).
    The H-Inception model is made of six Inception modules, where each module contains three convolution layers (in \colorbox{orangeF2}{orange}) and a MAxPooling layer (in \colorbox{magentaF2}{magenta}) followed by a concatenation (in \colorbox{yellowF2}{yellow}), a batch normalization layer (in \colorbox{oilF2}{oily}) and an activation function (in \colorbox{redF2}{red}).
    Each Inception module, except the first one, is proceeded by a bottleneck layer (in \colorbox{purpleF2}{purple}) to reduce the dimensionality and hence the number of parameters.
    The first Inception module contains the hybrid addition, which is the hand-crafted convolution filter (in \colorbox{green2F2}{green}).
    Residual connections exist between the input and the third module, as well as between the third module and the output (in \colorbox{cyanF2}{cyan}).}
    \label{fig:hinception_decompose}
\end{figure*}

Once the pre-trained model is fully trained, the model is then extended by a randomly initialized model.
The new constructed model, made of a pre-trained and a randomly initialized sub-model, is then fine tuned on the TSC task for each dataset independently (see Algorithm~\ref{alg:finetune}).
In summary, the different steps of the whole training procedure are:
\begin{itemize}
    \item Step 1: Given a set of $M$ UTSCD datasets:\\ $\{\mathcal{D}_0,\mathcal{D}_1,\ldots,\mathcal{D}_{M-1}\}$, where $\mathcal{D}_i\myeq\{(\textbf{x}_j,\textbf{y}_j)\}_{j=0}^{N_i-1}$, construct $\mathcal{D}_{PT} \myeq \{(\textbf{x}_n,\textbf{yd}_n)\}_{i=0}^{N-1}$, where $N\myeq\sum_{n=0}^{M-1}N_n$, is a dataset that includes all the time series samples from $\mathcal{D}_i$ with new labels $\textbf{yd}$ that represent the dataset the input sample $\textbf{x}$ belongs to.

    \item Step 2: Build a pre-trained model, $PT(.)$ with $L_{PT}$ layers trained on $\mathcal{D}$ to correctly classify the dataset each sample belongs to. See Algorithm~\ref{alg:pre-trained} for a detailed view on steps 1 and 2.

    \item Step 3: Build, for each of the $M$ datasets, a classifier $FT_i(.)$ for $i\in\{0,1,\ldots,M-1\}$ with $L_{PT}+L_{FT}$ layers.

    \item Step 4: Fine tune a classifier $FT_i(.)$ for each dataset. See Algorithm~\ref{alg:finetune} for a detailed view on steps 3 and 4.
\end{itemize}


\subsubsection{Backbone Model}
In this work, we base our model on the-state-of-the-art deep learning model for TSC in the literature, the Hybrid Inception architecture (H-Inception)~\cite{ismail-fawaz2022hccf}.
Its important to note that H-InceptionTime proposed in~\cite{ismail-fawaz2022hccf} is an ensemble of five H-Inception models trained with different initialization.
For this reason, the backbone architecture in our approach is the H-Inception architecture, and we ensemble the trained models as well following the original work~\cite{ismail-fawaz2022hccf,ismail2020inceptiontime}.
A summarized view of how the H-Inception backbone is decomposed into the pre-trained and fine tuning parts is presented in Figure~\ref{fig:hinception_decompose}.
Given that the original H-Inception architecture is made of six Inception modules, the first three modules are set to be part of the pre-trained model and the last three are then added to the fine tuning part.
We refer to our approach using this specific H-Inception backbone 
as PHIT (Pre-trained H-InceptionTime).

\subsubsection{Batch Normalization Multiplexer (BNM)}
\begin{figure}[!ht]
    \centering
    \includegraphics[width=0.7\linewidth]{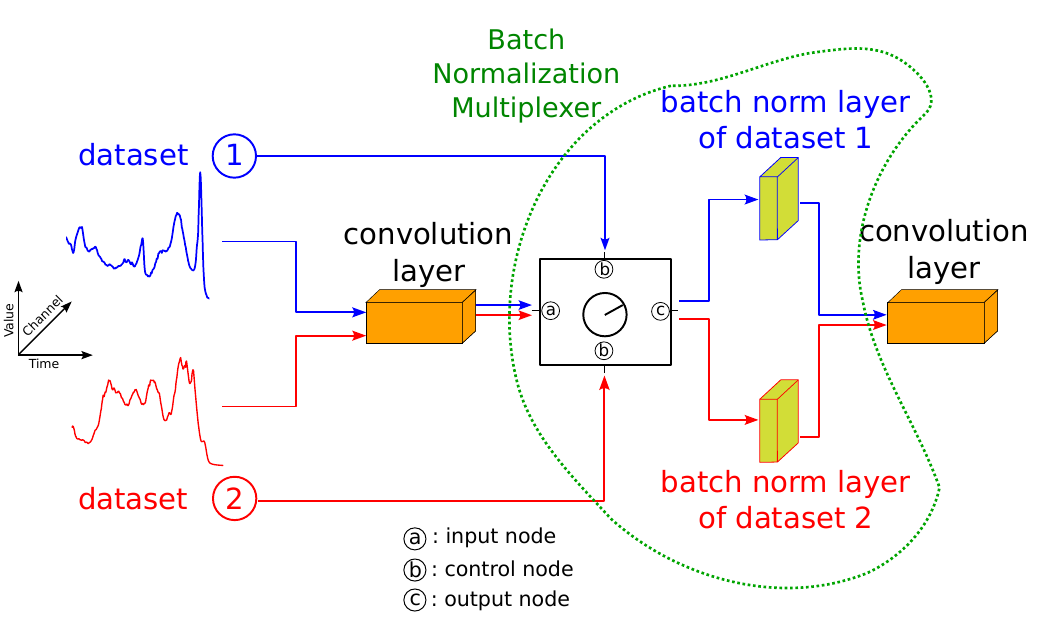}
    \caption{An example using the proposed Batch Normalizing Multiplexer (BNM) that solves the problem of learning a batch normalization layer on multiple samples of different distributions (datasets).
    The BNM is made of multiple batch normalization layers (in \colorbox{oilF2}{oily} with \colorbox{blueF2}{blue} and \colorbox{redF2}{red} contours) proceeded by a multiplexer.
    This multiplexer has three different nodes: (a) input node, where the input time series goes through, (b) the control node, where the information about the dataset this input time series belong to goes through, and (c) the output node.
    The path selected for the output node is controlled by the node (b).
    It is important to note that the BNM, such as the traditional batch normalization layer, learns on the whole batch.
    The only difference is that more than one batch normalization layer will be fed by parts of this batch, which intuitively means the flow of information is slower when using the BNM.}
    \label{fig:batch_norm}
\end{figure}

Most deep learning models for TSC~\cite{ismail2019deep} that achieve state of the art performance on the UCR archive~\cite{dau2019ucr} are convolution-based architectures that use the Batch Normalization layer with one of its goals is to accelerate the training.
In the backbone model we chose, H-Inception~\cite{ismail-fawaz2022hccf}, each convolution layer is followed by a Batch Normalization.
The role of the Batch Normalization is to learn how to scale and shift the batch samples in order to get a zero mean and unit variance.
This however may be problematic when samples in a same batch are generated from different distributions, in other words, from different datasets, such as in our pre-trained model's case.
For this reason, while training the pre-trained model on the pretext task, we should define multiple Batch Normalization layers for each dataset to replace the one batch normalization layer usually used in modern CNN architectures for TSC.
For this kind of layer to work, we should then give control to the model to connect the each sample in the batch to the correct batch normalization layer.
A visual representation of the proposed Batch Normalization Multiplexer (BNM) is presented in Figure~\ref{fig:batch_norm}.
From the figure, it can be observed the BNM takes as input the outcome of the previous layer, with the information of the dataset of the used series, this information is the same one the model is trying to predict.
The dataset information goes through the control node of the BNM and chooses which Batch Normalization layer the output node should be connected to.

\section{Results and Analysis}

\subsection{Experimental Setup}

\subsubsection{Datasets}
To evaluate the performance of our proposed approach, we conducted a series of experiments on the UCR archive dataset~\cite{dau2019ucr}, which comprises 128 datasets.
However, due to redundancies in the archive, our study narrows it down to only 88 datasets.
For instance, identical datasets appear multiple times but with varied train-test splits for distinct classification tasks.
Such overlaps could compromise the integrity of our model's training as it aims to predict the source dataset of a sample.
A scenario where identical series from two different datasets are included in the training set could confound the model's learning.
Moreover, some datasets, while seemingly distinct, merely had varied class counts or were truncated versions of another.
A detailed discussion of the reasons for excluding some datasets is reported in Appendix~\ref{sec:excluded_data}.
All datasets underwent a $z$-normalization prior to training to ensure a zero mean and unit variance.
As samples from these datasets may differ in length, zero padding was applied within each batch (rather than before training) to align with the length of the longest series.



\subsubsection{Division of the Datasets into Types}
The purpose of using a pre-trained model is that of boosting the performance of the deep learning classifier on small datasets using knowledge learned on large ones.
This is intuitively most applicable in the case where both the large and small datasets have at least basic information in common.
For this reason, we do eight different pretext experiments following the number of dataset types that exist in the UCR archive.
In other words, we used all of the datasets of the ECG type to train a pre-trained model and then fine tuned on each dataset independently.
These eight types with the corresponding number of datasets are the following:
\begin{itemize}
    \item Electrocardiogram (ECG): 7 datasets,
    \item Sensors: 18 datasets,
    \item Devices: 9 datasets,
    \item Simulation: 8 datasets,
    \item Spectrogram: 8 datasets,
    \item Motion: 13 datasets,
    \item Traffic: 2 datasets,
    \item Images contour: 23 datasets.
\end{itemize}

\subsubsection{Implementation Details}
The proposed method is implemented in \emph{Tensorflow python} and the code is publicly available~\footnote{https://github.com/MSD-IRIMAS/DomainFoundationModelsTSC}.
All of the parameters of the H-Inception model follow the same as in the original work~\cite{ismail-fawaz2022hccf}.
Each experiment was performed with five different initialization, including the pre-trained model and the fine tuned one.
Results of multiple runs were assembled together and the model used for evaluation is the best model monitored during training following the training loss.
We used a learning rate decay, ReduceLROnPlateau in \emph{keras}, to reduce the learning rate during training by monitoring the train loss with a factor of half.
All models were trained on a batch size of 64; the pre-trained model was trained for 750 epochs and the fine tuned model was trained for 750 epochs as well.
This last condition ensured us to not train the model for more epochs than the baseline (\ie, the baseline was trained for 1500 epochs following the original work~\cite{ismail-fawaz2022hccf}).
All experiments were conducted on a Ubuntu 22.04 machine with an NVIDIA GeForece RTX 3090 graphic card with a 24GB of memory.

\subsection{Comparing Pre-Training with Baseline}
\begin{figure}[!ht]
    \centering
    \includegraphics[width=0.45\linewidth]{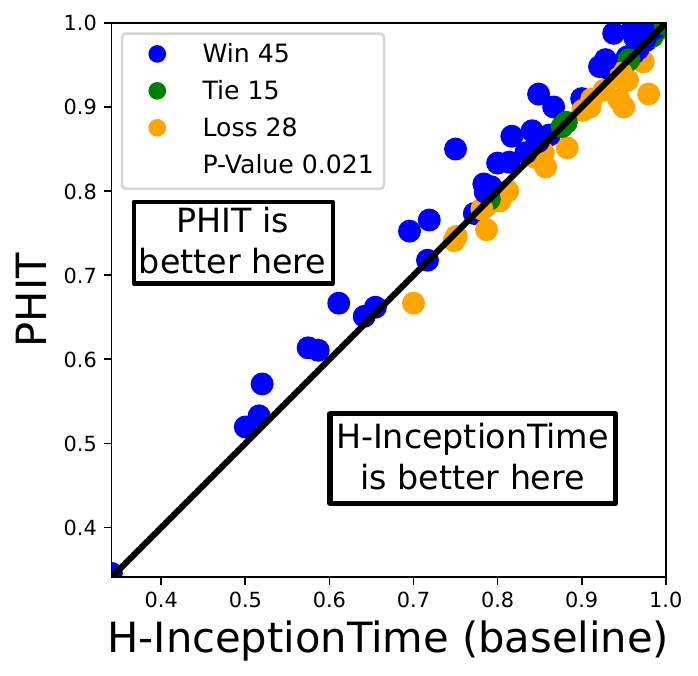}
    \caption{A 1v1 scatter plot that compares the performance of H-InceptionTime (baseline) and PHIT following the accuracy metric.
    Each point represents a dataset, where the $x$ and $y$ axis represent the accuracy of H-InceptionTime and PHIT, respectively.
    A blue point represents a win for PHIT, an orange point a win for H-InceptionTime and a green point a tie.}
    \label{fig:1v1_PHIT_baseline}
\end{figure}

We present in this section a 1v1 to compare our pre-training approach using H-Inception architecture to the baseline.
It is important to note that we compared the ensemble version of both the pre-training approach and the baseline.
We refer in what follows to our approach as Pre-Trained H-InceptionTime (PHIT).
Figure~\ref{fig:1v1_PHIT_baseline} represents this 1v1 comparison by a scatter plot between PHIT and H-InceptionTime.
Each point in this scatter plot represents a dataset of the UCR, where the $x$ and $y$ axis presents the accuracy metric of H-InceptionTime and PHIT, respectively.
The accuracy is evaluated on the test set for each dataset using both methods.
This 1v1 comparison resulted in concluding that over the 88 datasets, PHIT performs much better than the baseline.
From the legend of Figure~\ref{fig:1v1_PHIT_baseline} it can be seen that PHIT wins 48 times over the baseline; the baseline wins only 23 times.
To evaluate the statistical significance of this difference in performance, we presented as well a $p$-value produced using the Wilcoxon Signed-Rank Test.
This $p$-value, represents the $\%$ of confidence of a difference in performance being statistically significant.
If the $p$-value is less than $5\%$ it means there is not enough datasets to conclude a statistical significance in the difference of performance.
In this comparison, as seen in Figure~\ref{fig:1v1_PHIT_baseline}, the $p$-value between PHIT and the baseline is almost $0.09\%$, which means PHIT significantly outperforms the baseline.

\begin{table*}
\centering
\caption{The Win/Tie/Loss count between the proposed PHIT approach and the baseline (H-InceptionTime) per dataset domain.
The first column presents the number of datasets included per domain followed by the number of Wins for PHIT, number of Ties, and number of Wins for the baseline.
We include as well the percentage of number of losses and the average difference in accuracy (PHIT - baseline).
A positive value in the last column indicates that on average of all datasets in a specific domain, PHIT performs better than the baseline on the accuracy metric (lowest value 0.0 and highest value 1.0).}
\label{tab:wins_per_type}
\begin{tabular}{|c|c|c|c|c|c|c|}
\hline
\scriptsize
\textbf{\begin{tabular}[c]{@{}c@{}}Dataset\\ Type\end{tabular}} &
  \textbf{\begin{tabular}[c]{@{}c@{}}Number of\\ Datasets\end{tabular}} &
  \textbf{\begin{tabular}[c]{@{}c@{}}Wins of\\ PHIT\end{tabular}} &
  \textbf{\begin{tabular}[c]{@{}c@{}}Ties of\\ PHIT\end{tabular}} &
  \textbf{\begin{tabular}[c]{@{}c@{}}Losses of\\ PHIT\end{tabular}} &
  \begin{tabular}[c]{@{}c@{}}Percentage\\ of Losses\end{tabular} &
  \textbf{\begin{tabular}[c]{@{}c@{}}Difference in\\ Average Accuracy\\ (PHIT - Baseline)\end{tabular}} \\ \hline
\textbf{Devices}    & 9  & 4  & 0 & \textbf{5} & \textbf{55.55 \%} & +0.0046 \\ \hline
\textbf{ECG}        & 7  & \textbf{3}  & 2 & 2 & \textbf{28.57 \%} & +0.0012 \\ \hline
\textbf{Images}     & 23 & \textbf{14} & 2 & 7 & \textbf{30.43 \%} & +0.0087 \\ \hline
\textbf{Motion}     & 13 & \textbf{11} & 1 & 1 & \textbf{07.69 \%} & +0.0179 \\ \hline
\textbf{Sensors}    & 18 & \textbf{7}  & 5 & 6 & \textbf{33.33 \%} & +0.0002 \\ \hline
\textbf{Simulation} & 8  & \textbf{3}  & \textbf{3} & 2 & \textbf{25.00 \%} & +0.0051 \\ \hline
\textbf{Spectro}    & 8  & \textbf{3}  & 2 & \textbf{3} & \textbf{37.50 \%} & +0.0115 \\ \hline
\textbf{Traffic}    & 2  & 0  & 0 & 2 & \textbf{100.0 \%} & -0.0333 \\ \hline
\end{tabular}
\end{table*}

\subsubsection{Analysing Performance Per Domain}
In Table~\ref{tab:wins_per_type}, we present a detailed analysis on the performance of the proposed PHIT approach compared to the baseline per dataset domain.
We present, for each domain used in the UCR archive, the total number of datasets and the Win/Tie/Loss count with the average difference in performance in the last column.
A positive value in the last column confirms that on average PHIT outperforms the baseline on the average accuracy metric.
We also present in the 5th column the percentage of number of losses of PHIT.
From the table it can be seen that the percentage of losses never exceeds $50\%$ more than twice, and that the average difference in performance is always positive except on one type (\emph{Traffic}) where we only have two datasets.
These observations indicate that not only PHIT outperforms the baseline on a global scale of the UCR archive on the majority of domains.

This comparison shows that fine tuning a pre-trained model on a generic task which is in common between multiple datasets is significantly better than the traditional approach.
In what follows, we dig deeper into the cases in which the pre-trained model outperforms the baseline by studying the size of the training set.

\subsection{Larger Datasets Helping Smaller Datasets}\label{sec:large_help_small}

\begin{figure*}
    \centering
    \includegraphics[width=\textwidth]{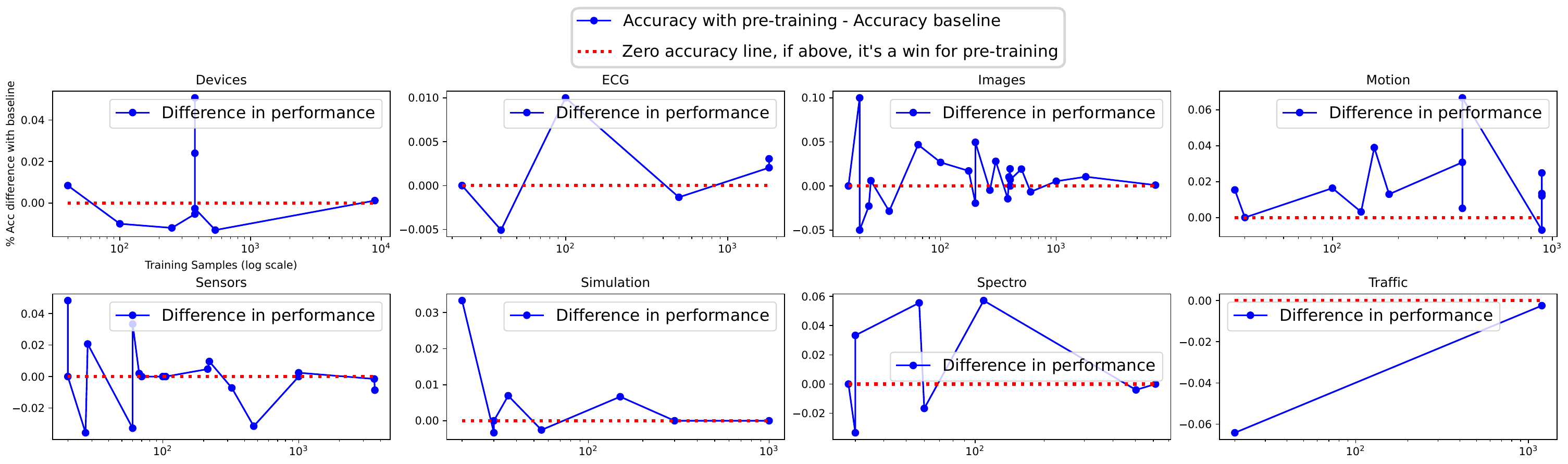}
    \caption{Comparing the performance of the proposed approach and its change with respect to the training set size.
    The curve represented in blue is the difference in performance between the proposed approach and the baseline.
    A positive value represents a win for the pre-training approach.
    For each plot, we show this comparison on the datasets of the same type in the UCR archive.
    The $x$-axis represents the number of training examples (in $\log_{10}$ scale).
    The $y$-axis represents the difference of accuracy between the usage of our pre-training approach and the baseline.}
    \label{fig:training_size}
\end{figure*}

This work is proposing a pretext task that consists of a pre-trained model to learn features on multiple datasets at the same time.
As detailed at the beginning, the purpose of this pretext task is to enhance the performance of deep learners on TSC tasks when the datasets presents very few number of training samples.
For this reason, we study in this section the effect of the pretext task on each of the 8 dataset types in function of the number of training samples.
This study is presented in Figure~\ref{fig:training_size}.
The figure represents the difference in accuracy between PHIT and the baseline on the $y$-axis and the training set size in $\log$ scale on the $x$-axis.
We present this study in 8 different plots, one for each type of dataset.
A positive value for the blue curves means a win for PHIT.
What can be observed in this study is that on average, the pretext task would boost datasets whose number of training samples is less than $10^3$ (on most examples and not all).
We argue this phenomena can be explained by considering the pretext task was able to extract knowledge more from larger datasets, while maintaining a transfer to the smaller ones.
This would eventually give the fine tuning stage a powerful information to learn the task of the small datasets and a bit of a noisy information to learn the task of the larger ones.
This is due to the fact that larger datasets are in need of full focus of the model on their own task.
This is not true with smaller datasets, where the model cannot learn alone with its own power: in this case it needs a push from a given source.

\subsection{Visualizing the Filters}
\begin{figure*}
    \centering
    \includegraphics[width=0.95\linewidth]{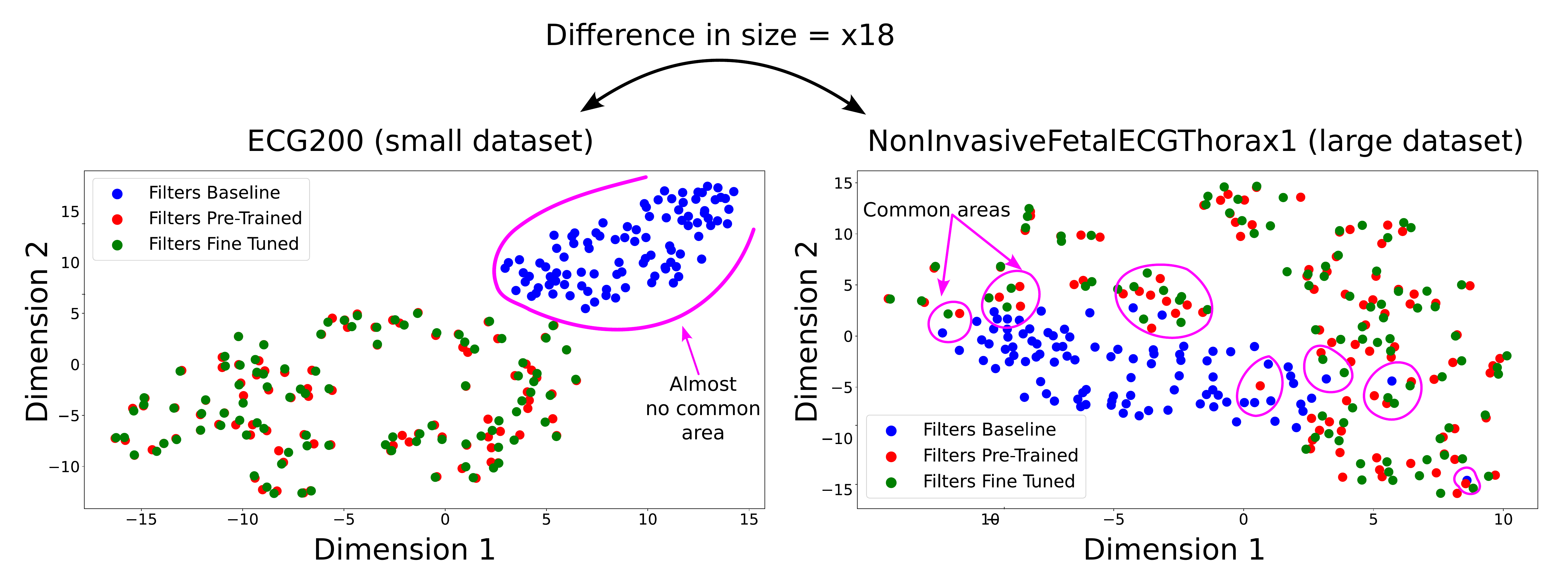}
    \caption{A two dimensional representation of the filters coming from the first Inception module of the baseline (in \colorbox{blueF2}{blue}), pre-trained(\colorbox{redF2}{red}) and fine tuned (\colorbox{greenF2}{green}) models.
    The used datasets in this study are ECG200 (left) and NonInvasiveFetalECGThorax1 (right).
    The two dimensional representation is done using t-SNE coupled with DTW to as a distance measure.
    The \colorbox{magentaF2}{magenta} areas represent the areas around the filters of the baseline model.}
    \label{fig:filters_ecg}
\end{figure*}

Since we base our work on CNNs, we can compare the space of the learned filters to see the effect of the pre-training approach.
In order to visualize this space, we used the t-Distributed Stochastic Neighbor Embedding (t-SNE)~\cite{van2008visualizing} visualization technique to reduce the dimensionality of the filters into a 2D plane.
The default usage of t-SNE is coupled with the Euclidean distance as a measure, but following the work in~\cite{ismail-fawaz2022hccf}, we used DTW instead.
This is due to the fact that convolution filters, such as time series, have an ordering dependencies between their elements and a shifted version of the filter is not a new one.
By taking the filters of the first Inception module from the baseline, the pre-trained model and the fine tuned model, we can visualize the filters in Figure~\ref{fig:filters_ecg}.
In this figure, we consider the experiment over the ECG datasets, where we choose a couple: ECG200 and NonInvasiveFetalECGThorax1.
The choice of these two datasets is not random: we chose these two given the difference in size of the training set.
For instance, ECG200 has $100$ training examples, whereas NonInvasiveFetalECGThorax1 has $1800$.

From Figure~\ref{fig:filters_ecg}, the filters of the baseline, pre-trained and fine tuned models are presented for each dataset.
The first noticeable aspect to see is that the blue points, representing the filters of the baseline, are quite different from the other red and green points.
This ensures that by using the pre-trained model then fine tune it, the backpropagation algorithm learns different convolution filters than the traditional baseline approach.
The second noticeable thing is that there exists a difference between both plots.
On the one hand, in the case of ECG200 (left plot), almost no common areas exist between the filters of the three models.
On the other hand, in the case of NonInvasiveFetalECGThorax1 (right plot) there exist many common areas between the filters of different colors.
This results in the same observation: we argued in Section~\ref{sec:large_help_small} that large datasets focus more on distilling knowledge rather than trying to find new features.
However, there exist some new areas for the pre-trained and fine tuned filters (green and red), which indicates that even though the dataset is large enough, the pre-trained model did explore new filters given what it learned from other datasets.


\begin{figure}[!ht]
    \centering
    \includegraphics[width=\linewidth]{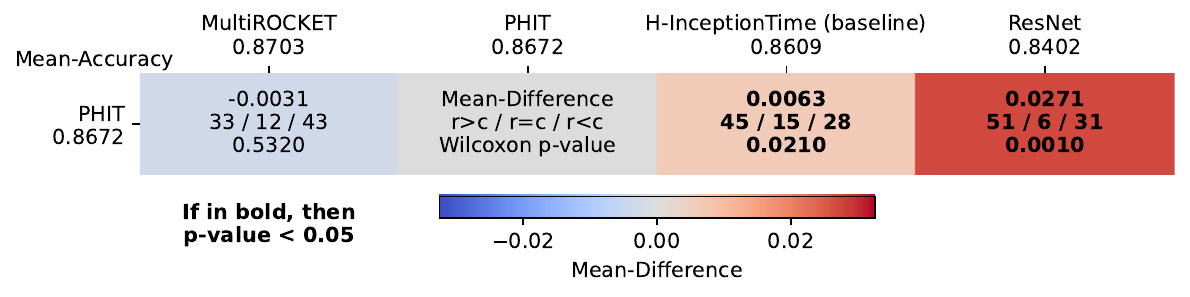}
    \caption{A Multi-Comparison Matrix (MCM) representing the comparison between the proposed approach PHIT
    with the state-of-the-art approaches.
    The winning approach following the average performance is MultiROCKET and in second comes our approach.
    No conclusion can be found on the difference of performance between MultiROCKET and PHIT given the high $p$-value.}
    \label{fig:mcm_sota_PHIT}
\end{figure}

\subsection{Comparison with the State-of-the-Art}

In what follows, we utilize a comparison technique proposed in~\cite{ismail2023approach} called the Multi-Comparison Matrix (MCM).
This MCM presents a pairwise comparison between the classifiers as well as their ordering following the average performance.
The MCM has shown to be stable to the addition and removal of classifiers, which gives it an advantage over other comparison approaches.
The MCM presents as well the Win/Tie/Loss count and a $p-$value generated using the two tailed Wilcoxon Signed-Ranked Test to study the significance in the difference of performance.
The MCM presents as well an ordering of performance of all classifiers following their average performance.
In what follows, we present the MCM to compare PHIT to the state-of-the-art approaches including deep and non-deep learning approaches in Figure~\ref{fig:mcm_sota_PHIT}.
It can be concluded that on the 88 datasets of the UCR archive, PHIT outperforms all of the deep learning approaches following the average performance metric.
The MCM also shows that given the 88 datasets, no conclusion can be found on the statistical significance difference in performance between PHIT and the state-of-the-art MultiROCKET.

\begin{figure}[!ht]
    \centering
    \includegraphics[width=0.8\linewidth]{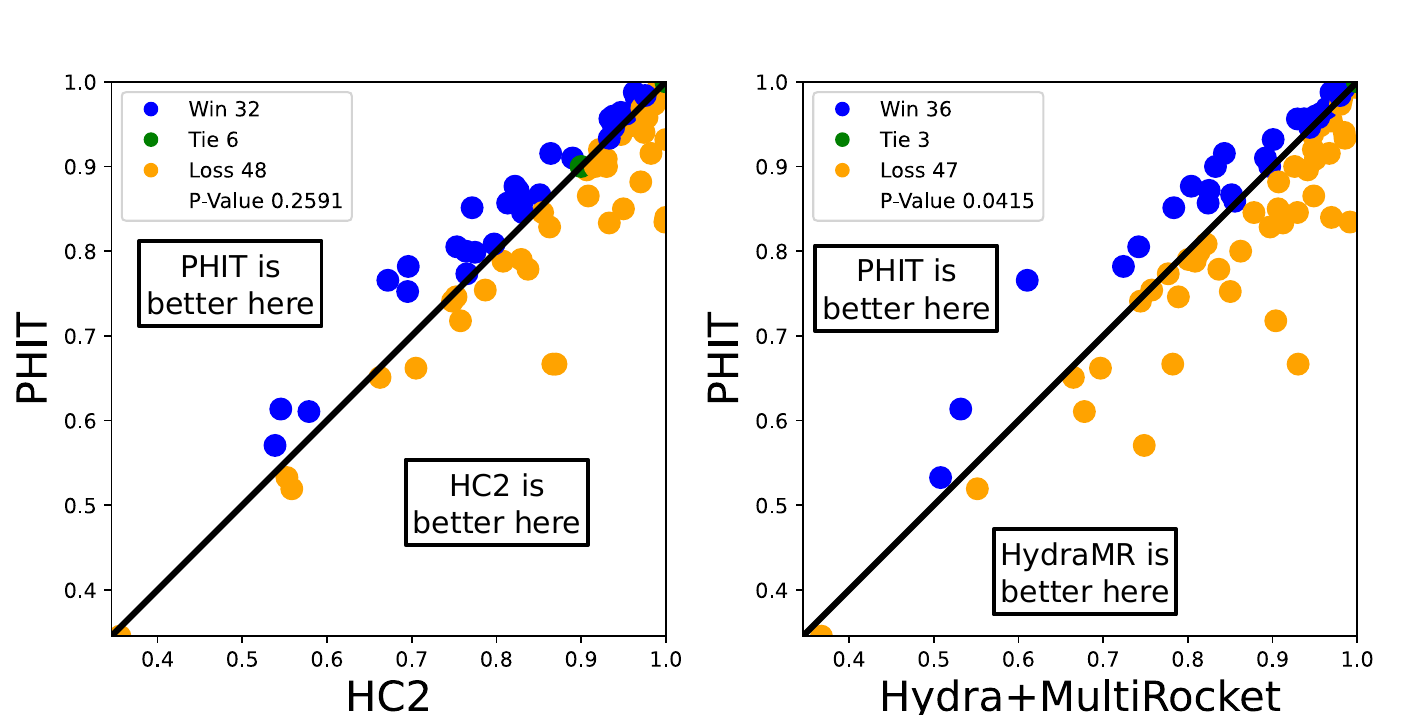}
    \caption{Two 1v1 scatter plots representing the comparison between the proposed approach, PHIT, with two state-of-the-art models for TSC, HIVE-COTE2.0 (HC2) and HydraMultiROCKET (HydraMR).}
    \label{fig:1v1_hc2_hydraMR}
\end{figure}

In order to also compare our approach with HIVE-COTE2.0 (HC2)~\cite{middlehurst2021hive} and Hydra+MultiROCKET (HydraMR)~\cite{dempster2023hydra,middlehurst2023bake}, we only used 86 datasets given that for some datasets of the UCR archive the results are not provided on the original versions for these two models.
The scatter plots showing the performance of PHIT compared to HC2 and HydraMR are presented in Figure~\ref{fig:1v1_hc2_hydraMR}.
On one hand, this figure shows that PHIT is still not as good as the HydraMR though the scatter plot shows that on 34 datasets, PHIT wins with a significant margin.
On the other hand, no conclusion can be made on the statistical significance in the difference of performance between HC2 and PHIT.
This concludes that the proposed approach is able to boost a lot the baseline deep learner to achieve HC2 state-of-the-art performance.


\section{Conclusions}
In this work, we addressed the Time Series Classification problem by employing innovative pre-trained domain foundation models effectively mitigating overfitting issues in small datasets. Leveraging the UCR archive for evaluation, our methodology involved training models on multiple datasets to accurately classify each sample’s original dataset. Subsequent fine-tuning of these models on individual datasets demonstrated superior performance over traditional methods, as evidenced by comprehensive experiments and analyses on the UCR datasets. Our contribution is the creation of domain-specific pre-trained foundation models for time series datasets in the UCR archive, offering a resource for researchers and paving the way for future extensions. This approach, with its inherent generic filters, holds promise for efficient adaptation to new datasets, potentially revolutionizing the training process in time series classification.

\section*{Acknowledgment}
This work was supported by the ANR DELEGATION project (grant ANR-21-CE23-0014) of the French Agence Nationale de la Recherche. The authors would like to acknowledge the High Performance Computing Center of the University of Strasbourg for supporting this work by providing scientific support and access to computing resources. Part of the computing resources were funded by the Equipex Equip@Meso project (Programme Investissements d’Avenir) and the CPER Alsacalcul/Big Data. The authors would also like to thank the creators and
providers of the UCR Archive.

\bibliographystyle{splncs04}
\bibliography{biblio}

\appendix
\section{Excluded Datasets}\label{sec:excluded_data}
In Table~\ref{tab:excluded_datasets}, we presents all of the excluded datasets from this study with the reason for their exclusion.
\begin{table*}
\centering
\caption{Excluded datasets from the UCR archive in this study.
Each dataaset is followed by its information and a reason for its exclusion.
The authors would like to thank the maintainer of the websites \text{https://www.timeseriesclassification.com}  and \text{https://www.cs.ucr.edu/\~{}eamonn/time\_series\_data\_2018/} from which the information presented in this table were gathered.}
\label{tab:excluded_datasets}
\scriptsize
\begin{tabular}{|c|c|c|c|c|c|}
\hline
\textbf{Type} &
  \textbf{Dataset} &
  \textbf{Train Samples} &
  \textbf{Test Samples} &
  \textbf{Length} &
  \textbf{reason (if excluded)} \\ \hline
EOG &
  EOGHorizontalSignal &
  362 &
  362 &
  1250 &
  \begin{tabular}[c]{@{}c@{}}same datasets multivariate\\ with 2 channels divided\\ into 2 univariate datasets\end{tabular} \\ \cline{2-5}
 &
  EOGVerticalSignal &
  362 &
  362 &
  1250 &
   \\ \hline
EPG &
  InsectEPGRegularTrain &
  62 &
  249 &
  601 &
  \begin{tabular}[c]{@{}c@{}}same test set, different\\ train set size, a combination\\ of both train sets is better\\ than doing a pretext task\end{tabular} \\ \cline{2-5}
 &
  InsectEPGSmallTrain &
  17 &
  249 &
  601 &
   \\ \hline
 &
  PigAirwayPressure &
  104 &
  208 &
  2000 &
   \\ \cline{2-5}
Hemodynamics &
  PigArtPressure &
  104 &
  208 &
  2000 &
  \begin{tabular}[c]{@{}c@{}}correlation unclear between\\ these three datasets\end{tabular} \\ \cline{2-5}
 &
  PigCVP &
  104 &
  208 &
  2000 &
   \\ \hline
HRM &
  Fungi &
  18 &
  186 &
  201 &
  Only one dataset in this type \\ \hline
 &
  DistalPhalanxOutlineAgeGroup &
  400 &
  139 &
  80 &
  \begin{tabular}[c]{@{}c@{}}same samples as DistalPhalanxTW\\ with different classification and\\ train test split\end{tabular} \\ \cline{2-5}
 &
  DistalPhalanxOutlineCorrect &
  600 &
  276 &
  80 &
   \\ \cline{2-6} 
 &
  FaceAll &
  560 &
  1690 &
  131 &
  \begin{tabular}[c]{@{}c@{}}same as FacesUCR with different\\ train test split\end{tabular} \\ \cline{2-6} 
 &
  FiftyWords &
  450 &
  455 &
  270 &
  \begin{tabular}[c]{@{}c@{}}same as WordSynonyms with\\ more classes\end{tabular} \\ \cline{2-6} 
Image &
  MiddlePhalanxOutlineAgeGroup &
  400 &
  154 &
  80 &
  Same reason as DistalPhalanx \\ \cline{2-5}
 &
  MiddlePhalanxOutlineCorrect &
  600 &
  291 &
  80 &
   \\ \cline{2-6} 
 &
  ProximalPhalanxOutlineAgeGroup &
  400 &
  205 &
  80 &
  Same reason as DistalPhalanx \\ \cline{2-5}
 &
  ProximalPhalanxOutlineCorrect &
  600 &
  291 &
  80 &
   \\ \cline{2-6} 
 &
  MixedShapesRegularTrain &
  500 &
  2425 &
  1024 &
  \begin{tabular}[c]{@{}c@{}}Bigger version of\\ MixedShapesSmallTrain\end{tabular} \\ \cline{2-6} \hline
 &
  GunPoint &
  50 &
  150 &
  150 &
  \begin{tabular}[c]{@{}c@{}}GunPointAgeSpan is the new\\ version with more samples\end{tabular} \\ \cline{2-6} 
 Motion &
  WormsTwoClass &
  181 &
  77 &
  900 &
  \begin{tabular}[c]{@{}c@{}}Same as Worms with different\\ number of classes\end{tabular} \\ \cline{2-6}
 &
  GunPointMaleVersusFemale &
  135 &
  316 &
  150 &
  \begin{tabular}[c]{@{}c@{}}Same as AgeSpan version with\\ different train test split\end{tabular} \\ \cline{2-5}
 &
  GunPointOldVersusYoung &
  136 &
  315 &
  150 &
   \\ \hline
Power &
  PowerCons &
  180 &
  180 &
  144 &
  Only one dataset for this type \\ \hline
 &
  AllGestureWiimoteX &
  300 &
  700 &
  Vary &
   \\ \cline{2-5}
 &
  AllGestureWiimoteY &
  300 &
  700 &
  Vary &
  \begin{tabular}[c]{@{}c@{}}too much Variable length datasets\\ to handle in this type for the pretext\\ task which already has the variable\\ length issue/instability\end{tabular} \\ \cline{2-5}
 &
  AllGestureWiimoteZ &
  300 &
  700 &
  Vary &
   \\ \cline{2-6} 
 &
  DodgerLoopDay &
  78 &
  80 &
  288 &
   \\ \cline{2-5}
 &
  DodgerLoopGame &
  20 &
  138 &
  288 &
  \begin{tabular}[c]{@{}c@{}}All dodger datasets are the same\\ with different train test split\\ with too many missing values\end{tabular} \\ \cline{2-5}
Sensors &
  DodgerLoopWeekend &
  20 &
  138 &
  288 &
   \\ \cline{2-6} 
 &
  FreezerRegularTrain &
  150 &
  2850 &
  301 &
  \begin{tabular}[c]{@{}c@{}}Same as FreezerSmallTrain with\\ more training examples\end{tabular} \\ \cline{2-6} 
 &
  GesturePebbleZ1 &
  132 &
  172 &
  Vary &
   \\ \cline{2-5}
 &
  GesturePebbleZ2 &
  146 &
  158 &
  Vary &
  \begin{tabular}[c]{@{}c@{}}too much Variable length datasets\\ to handle in this type for the pretext\\ task which already has the variable\\ length issue/instability\end{tabular} \\ \cline{2-5}
 &
  PickupGestureWiimoteZ &
  50 &
  50 &
  Vary &
   \\ \cline{2-5}
 &
  ShakeGestureWiimoteZ &
  50 &
  50 &
  Vary &
   \\ \hline
 &
  Rock &
  20 &
  50 &
  2844 &
  Not a time series \\ \cline{2-6} 
 &
  SemgHandGenderCh2 &
  300 &
  600 &
  1500 &
   \\ \cline{2-5}
Spectrum &
  SemgHandMovementCh2 &
  450 &
  450 &
  1500 &
  \begin{tabular}[c]{@{}c@{}}Same datasets different split\\ if we include one of them we end up\\ with one dataset for this type\end{tabular} \\ \cline{2-5}
 &
  SemgHandSubjectCh2 &
  450 &
  450 &
  1500 &
   \\ \hline
 &
  GestureMidAirD1 &
  208 &
  130 &
  Vary &
   \\ \cline{2-5}
Trajectory &
  GestureMidAirD2 &
  208 &
  130 &
  Vary &
  \begin{tabular}[c]{@{}c@{}}Only datasets of variable length.\\ The three datasets are from the same\\ distribution of a 3D multivariate\\ time series with each being a dimension\\ a more suitable approach is to combine\\ the three datasets and solve\\ a multivariate TSC task\end{tabular} \\ \cline{2-5}
 &
  GestureMidAirD3 &
  208 &
  130 &
  Vary &
   \\ \hline
\end{tabular}
\end{table*}

\end{document}